\documentclass[12pt, fullpage, conference]{IEEEtran}
\IEEEoverridecommandlockouts
\usepackage{cite}
\usepackage{amsmath,amssymb,amsfonts}
\usepackage{algorithmic}
\usepackage{graphicx}
\usepackage{textcomp}
\usepackage{xcolor}
\usepackage{verbatim}
\usepackage{geometry}
\def\BibTeX{{\rm B\kern-.05em{\sc i\kern-.025em b}\kern-.08em
    T\kern-.1667em\lower.7ex\hbox{E}\kern-.125emX}}
    
\begin{document}

\title{Classifying Video based on Automatic Content Detection Overview}

\author{\IEEEauthorblockN{Yilin Wang}
\IEEEauthorblockA{\textit{Computer Science in Multimedia} \\
\textit{University of Alberta}\\
Edmonton, Canada \\
yilin28@alberta.ca}
\and
\IEEEauthorblockN{Jiayi Ye}
\IEEEauthorblockA{\textit{Computer Science in Multimedia} \\
\textit{University of Alberta}\\
Edmonton, Canada \\
jye8@alberta.ca}

}

\maketitle

\begin{abstract}
Video classification and analysis is always a popular and challenging field in computer vision. It is more than just simple image classification due to the correlation with respect to the semantic contents of subsequent frames brings difficulties for video analysis. In this literature review, we summarized some state-of-the-art methods for multi-label video classification. Our goal is first to experimentally research the current widely used architectures, and then to develop a method to deal with the sequential data of frames and perform multi-label classification based on automatic content detection of video. We will use a large-scale dataset --- the Youtube-8M dataset ~\cite{abu2016youtube} to train and evaluate our proposed method.
\end{abstract}


\section{Introduction}
\label{intro}
With the development of Internet technologies, a large number of videos are uploaded to the network every day, thus how to understand and efficiently use these video content has become a great challenge. It also has a wide range of application scenarios such as video recommendation, video search and so on.
Video classification refers to the classification of the content contained in a given video clip. Although video contains not only image information, but also audio information, for the vast majority of videos , the image information can reflect the video content more directly. We mainly study video classification based on vision, that is, we mainly focus on the spatial and temporal characteristics of the video itself, while audio information only plays a supplementary role. As a result, some vision-based methods for object tracking or motion detection can also be used, for instance, \cite{ELNAGAR19951537} proposed a motion detection methods using background constrains, pose recognition methods proposed by \cite{inproceedings} and  methods for exploiting 3D human data \cite{berretti2018representation}, can also aid in analysis in video content.

In this literature review, we summarized some state-of-the-art methods for multi-label video classification. In section 2, We first briefly describe the data format commonly used in video classification tasks, also pointing out some famous datasets. Then we generalize some proposed methods of preprocessing. In section 3, we introduce techniques related to video content classification. In section 4, we listed some evaluation methods. Finally, we discuss video classification topics in broad outline in section 5.

\section{Data set processing and transmission}
Video classification/behavior recognition is a very challenging topic in the field of computer vision, because it not only needs to analyze the spatial information of the target body, but also needs to analyze the information in the time dimension. A good action recognition problem data set should have enough variety of frames and theme types to generalize the trained architecture into many different tasks.

\subsection{Dataset: YouTube-8M}

YouTube-8M is one of the largest multi-label video classification dataset, composed of 8 million videos. It is a benchmark dataset for video understanding, where the main task is to determine the key topical themes of a video. This dataset represent samples as a sequence of vectors , each vector is an embedded representation of a frame and an audio piece. The visual and audio features are pre-extracted with Inception Net and provided with the dataset. Youtube-8M uses Knowledge Graph entities to succinctly describe the main themes of a video.There are 4800 Knowledge Graph entities in total, organized into 24 top-level verticals. Each entity contains at least 200 video samples.

\subsection{Dataset:Activity Net}

Activity Net\cite{caba2015activitynet} is a large-scale video benchmark for human activity understanding. The dataset contains more than 648 hours of untrimmed video, totaling about 20,000 videos. It contains 200 different daily activities such as' walking the dog ', 'long jump', and vacuuming floor. It is the largest motion analysis video data set, including two tasks of classification and detection.
The current version of Activity Net Dataset is V1.3, including 20,000 YouTube videos (the training set contains about 10,000 videos, and the verification set and the test set each contain about 5,000 videos), a total of about 700 hours of videos, and each video has 1.5 action instances on average.

\begin{figure}[htbp]
\centerline{\includegraphics[width=3in]{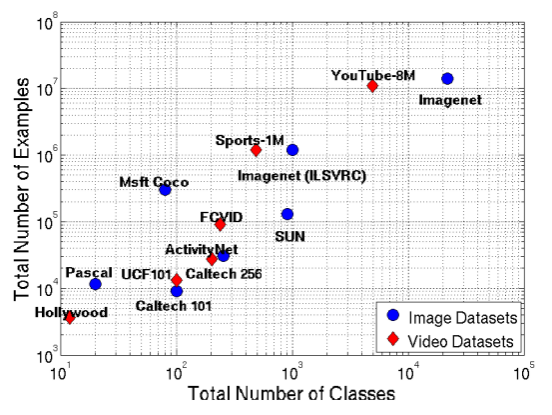}}
\caption{A comparison between different video and image classification datasets on scale.}
\label{compare}
\end{figure}

\subsection{Data Compression and transmission}

With the development of science and technology, the forms of expression of media are becoming more and more diversified, and the types of videos are gradually increasing, such as panoramic video\cite{baldwin1999panoramic}. How to transmit video data efficiently has become a hot issue\cite{basu1993variable}. \cite{basu1998enhancing} a method that considered human vision as a variable resolution system. The method of compression accords with the visual habits of the person. \cite{basu1994videoconferencing}improved this method by combining information obtained from multiple points-of-interest (foveae) in an image. In addition to the compression method based on human visual habits, a motion capture based algorithm commonly used in 3D data can also be used in video compression\cite{firouzmanesh2011perceptually}. 

In the problem of video classification, to achieve more accurate action or emotion classification, sometimes human face recognition is needed to assist \cite{yin1999integrating}. Among the facial recognition problems, the nose is the most difficult to locate among the five tubes. However, \cite{yin2001nose} solves this problem through a feature detection method on the facial organ areas, while \cite{bernogger1998eye} gets a new compression method through eye tracking. 

\section{Video Content Classification}
Complex video classification tasks often have more than one label, so many video classification tasks can also be considered as multi-label classification tasks. Here are some common methods of video classification and multi-label classification.

\subsection{Bag of visual words}

The Bag of Words (BoW) model is the simplest form of text representation in numbers. In other words, a sentence can be represents as a bag of words vector. The Bag of Words create vectors containing the frequency of words in the document. The standards steps first proposed by \cite{1238663} described the process, parsing the documents into words firstly, and represents them as stems, finally using a stop list to filter out common words.The remaining words are assigned a unique identifier to construct vectors.

However, if the sentences' lengths are very different, this representation may introduces many 0s, resulting a sparse matrix. The memory required to store vectors will also increase. Moreover, by storing words separately in this way, we lose the correlated information between words. One improvement is to apply weighting to the components of this vectors. The Term Frequency–Inverse Document Frequency method is known as the standard weighting methods for word vectors.

Bag of visual words (BOVW) borrow the idea from Bag of Words,  is now commonly used in many  classification problems. In this context, image features are used  as the “words”. Features contains keypoints and descriptors. Constructing BOVW  usually use feature extractor such as SIFT detect features and extracting descriptors.

\begin{figure}[htbp]
\centerline{\includegraphics[width=3in]{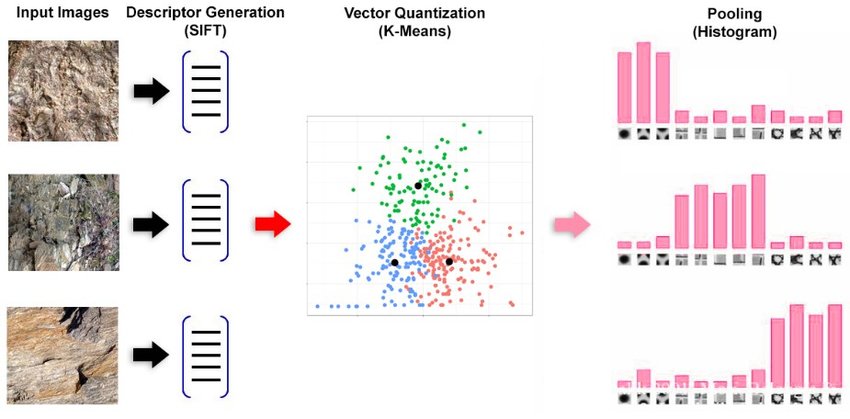}}
\caption{Bag of visual words, figure from ~\cite{inbook1}}
\label{fig}
\end{figure}

\subsection{Mixture of expert}
Mixture of expert (MOE) is a commonly used model for classification problems. It is a kind of neural network combines multiple models. There are different ways of generating data that are appropriate for a dataset. Different from the general neural network, it trains multiple models separately according to the data. Each model is called an expert, and the gating module is used to select which expert to use. The actual output of the model is the weight combination of the output of each model and the gating model. Different functions (linear or nonlinear) can be used for each expert model. An Mixture of expert system is the integration of multiple models into a single task.
\cite{wang2017monkeytyping}and \cite{kimgoogle} both used this method to solve the problem of multi-label video classification.

\subsection{Classifier Chains}

The basic idea of this algorithm is to transform the multi-label learning 
problem into a chain of binary classification problems, where subsequent binary classifiers in 
the chain is built upon the predictions of preceding ones\cite{zhang2013review}. In a classifier chain, each classifier use both input feature and the results of previous models to predict a single label. \cite{wang2017monkeytyping}is one of the model that uses this method. The whole architecture of chaining in this model consists of several stages. Each chaining unit in one stage accepts a feature vector, either from input directly or from a representation of other sub-model, and all the predictions from earlier stages.

\subsection{Learnable pooling with Context Gating for
video classification}
\label{classification}

General methods for video analysis usually extract features from the interval frames of a video, and use averaging or maximum pooling for feature aggregation over time. However, methods like LSTM is easy to fall into a local optimal solution. As a result, ~\cite{miech2017learnable} proposed a non-recurrent
aggregation techniques for aggregating features. They also introduce an efficient non-linear unit Context Gating layer to re-weight the features and the output labels.

\begin{figure}[htbp]
\centerline{\includegraphics[width=3in]{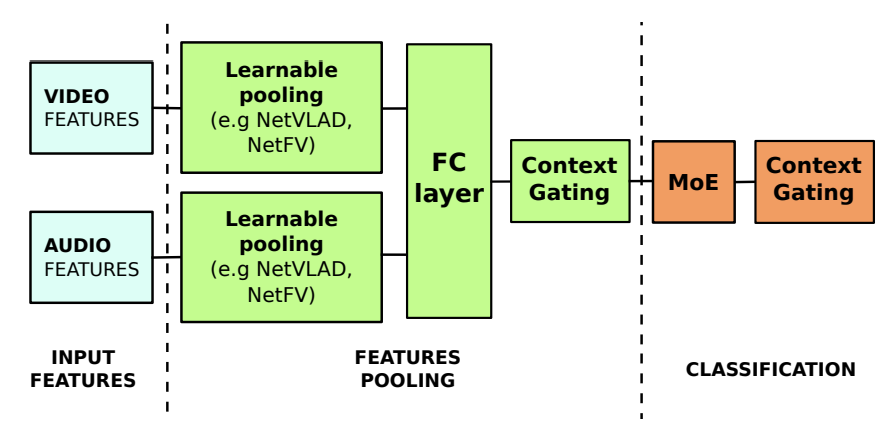}}
\caption{Architecture of ~\cite{miech2017learnable}.}
\label{fig}
\end{figure}

Models are trained using the Adam algorithm with the cross entropy loss and implemented by the TensorFlow framework. The results shown in their paper outperform over many recurrent models.

In the 2nd YouTube-8M video understanding challenge, ~\cite{Skalic_2018_ECCV_Workshops} further improve the previous proposed work and ensemble submodels belonging to Fisher vectors, NetVlad, Deep Bag of Frames and Recurrent
neural networks to achieve efficient classification under model size constrains.

\section{Evaluation}

\subsection{Mean Average Precision}
There is more than one label on the each element in multi-label image classification, so the standard of common single-label classification, namely mean accuracy, cannot be used for evaluation. This task adopts a method similar to that used in information retrieval -- MAP (Mean Average Precision). In the Youtube-8M dataset the evaluation function is as below:
\begin{equation}
R =\frac{T_{p}} {T_{p}+F_{n}} 
\end{equation}
\begin{equation}
 P =\frac{T_{p}} {T_{p}+F_{p}}
\end{equation}

R represents the recall rate and P is the precision.

True positives(TP): The number of positive examples that are correctly classified, i.e. the number of instances that are actually positive examples and are classified as positive examples by the classifier.	

False positives(FP): The number of incorrectly classified as positive examples, i.e. the number of instances that are actually negative but are classified as positive by the classifier.

False negatives(FN): the number of instances that are wrongly classified as negative examples, i.e., the number of instances that are actually positive examples but are classified as negative examples by the classifier;

True negatives(TN): the number of instances correctly classified as negative cases, i.e., the number of instances that are actually negative cases and classified as negative cases by the classifier.

The average precision, approximating the area under the precision-recall curve, can then be
computed as follow, where$\tau _{j} =\frac{j}{10000}$:
\begin{equation}
AP=\sum_{j=1}^{10000}P(\tau_{j})[R(\tau_{j})-R(\tau_{j+1})]
\end{equation}

\subsection{Precision at equal recall rate }
In the Youtube-8M dataset, the video level annotation precision can be measured when retrieving the same number of entities per video as there are in the ground-truth. PERR can be written as:
\begin{equation}
\frac{1}{\left | V:\left | G_{v}  \right | >0 \right | } \sum_{\nu\in  V:\left | G_{v}  \right | >0}^{}\left [ \frac{1}{G_{v}} \sum_{e\in G_{v} }^{}\mathbb{I}(rank_{_{v,e}}\le \left | G_{v}  \right |  ) \right ]  
\end{equation}

\section{Conclusion}
In conclusion, video classification is a challenging task. It includes feature extraction and aggregation, 

In the feature extraction process, changes in viewpoint, illumination and partial occlusion can greatly effect the information retrieval results of general methods.  
Although many noise reducing methods have been used to filter out unstable regions, however, it may be suboptimal.

For feature aggregation, traditional methods aggregating information over a sequence of frames like RNNs requires large scale data and are time-consuming. Moreover, some assumptions have been made during the process, for example, assuming similarity between frames, which is not always true.

In this literature review, we analyze some state-of-the-art methods for video classification, in order to provide a foundation for our project.

\section{References}


\bibliographystyle{IEEEtran}
\bibliography{egbib}

\end{document}